# Can a Frontier LLM Repair a Quantized Network by Tool-Free Weight-Space Programming?

Steffen Urban
Independent Researcher

We give four large language models (LLMs) of the Claude family the complete weights of a 20k-parameter CIFAR-10 CNN as text and ask them to produce a 4-bit network without tools, data or the ability to run inference. Standard round-to-nearest (RTN) quantization to 4 bit lowers the test accuracy from 81.4% to 60.4%. We design four evaluation protocols that differ in the input and task description given to the LLMs and analyze their responses. In the first two protocols the LLMs are given the float and the 4-bit RTN network and are told that the accuracy has dropped significantly. In the last two they receive only the float network and the task to quantize it, which tests whether they apply an optimization on their own or fall back to plain RTN. Our analysis shows that a frontier LLM can recover most of the lost accuracy. Given one answer per layer, Opus 5.5 raises the accuracy of the 4-bit network to 77.95% by balancing the rounding errors within each 3×3 kernel, reasoning over the integer weights without tools. This is 3.5 pp below the float network and 2.8 pp below AdaRound, which learns the rounding from calibration data. Given only the float network, the LLMs name error-cancelling corrections but do not apply them and fall back to plain RTN, at most with mildly clipped scales, which still lifts Fable 5.2's quantized network to 70.0%. We analyze in detail what the LLMs did and why it works. Code and prompts can be found at `github.com/urbste/weight-space-programming-ptq`.

## 1 Introduction

Post-training quantization (PTQ) converts a trained float network to low-bit integers without retraining [1], [2], [3]. Its simplest form, per-channel symmetric round-to-nearest (RTN), is close to lossless at 8 bit for most CNNs, while at 4 bit its accuracy loss varies widely between networks and can be large [2], [3]. Better PTQ methods fix this with calibration data and layer-wise reconstruction [4], [5], [6]. Data-free methods [7], [8], [9] work from the checkpoint alone, but they too are fixed numerical procedures.

We ask whether a large language model (LLM) can find and execute such a repair by *reading* the weights. The LLM gets the weights of a small CNN as text, either the float weights alone or together with the RTN integer weights, and writes integer weights and scales. It has no code interpreter, no calculator, no data and cannot evaluate the network. All arithmetic happens in its own reasoning. The setting is deliberately extreme. Of course, a short Python script quantizes better and faster. The question is what the LLM notices in the numbers, how accurately it executes what it decides, and what limits it.

To this end, we train a simple 20k-weight CNN on CIFAR-10 [10] that will fit into the context of current LLMs, which makes the question testable. We evaluate **four protocols** (Section 3.2) that differ in whether the LLM sees the RTN network and is told about the accuracy drop, whether it gets BatchNorm statistics, and whether it answers in one response or in one response per chunk.

We find that frontier LLMs without tools arrive at a published data-free PTQ correction by reading raw weights. Opus 5.5 and Sonnet 5.5 independently balance the per-kernel rounding error sums (kernel balancing, Section 4.2.1), the kernel-wise step of SQuant [9]. With one answer per chunk, Opus 5.5 applies it to all 2,200 kernels by hand (96% of edits identical to exact kernel balancing) and reaches 77.95% (Section 4.2.2), so the one-answer result is limited by the amount of arithmetic per answer. A first-order error model shows that the per-kernel error sum sets the mean shift of each output, which explains why the rule works (Section 5.1). Given only the float weights, the LLMs write RTN, at most with clipped outlier rows, and none uses the BatchNorm statistics to cancel the mean shift (Section 4.2.3, Section 5.2).

## 2 Related work

*Post-training quantization.* Per-channel RTN with BatchNorm folding is the standard baseline [1], [2]. Clipping the range trades clipping error for resolution, analytically in ACIQ [11] and by MSE search in most toolchains [3]. Outlier channel splitting [12] avoids clipping by duplicating channels. Rounding to nearest is not optimal for the layer output. AdaRound [4] learns the rounding direction per weight from calibration data, BRECQ [5] reconstructs blocks, and GPTQ [6] uses second-order information to compensate errors column by column. Without data, DFQ [7] equalizes ranges across layers and corrects the bias shift caused by rounding, estimating input statistics from BatchNorm parameters. ZeroQ [8] synthesizes calibration data from BatchNorm statistics. SQuant [9] approximates the Hessian by element-, kernel- and channel-wise terms and flips rounding decisions until the absolute rounding-error sum of every kernel is at most half a step, then does the same per output channel while allowing one step per kernel. Vendor toolchains for edge accelerators differ in scaling and clipping, so the same checkpoint can reach different accuracy on different backends [13]. The correction our LLMs find, constraining the sum of rounding errors per kernel, is the kernel step of SQuant. AdaRound learns rounding decisions of this kind from data.

*LLMs over numbers and weights.* LLMs can act as optimizers that propose solutions from a history of scored attempts [14], or drive program search with an evaluator in the loop [15]. Both rely on feedback, which we remove. Weight-space learning trains networks that take other networks' weights as input [16], [17]. We instead ask a general LLM to reason over weights in context. Bills et al. [18] use LLMs to explain individual neurons from activations, while here the LLM sees only parameters. Weight-space programming [19] lets LLM agents edit network weights with tools and evaluation. Our setting is its tool-free limit,

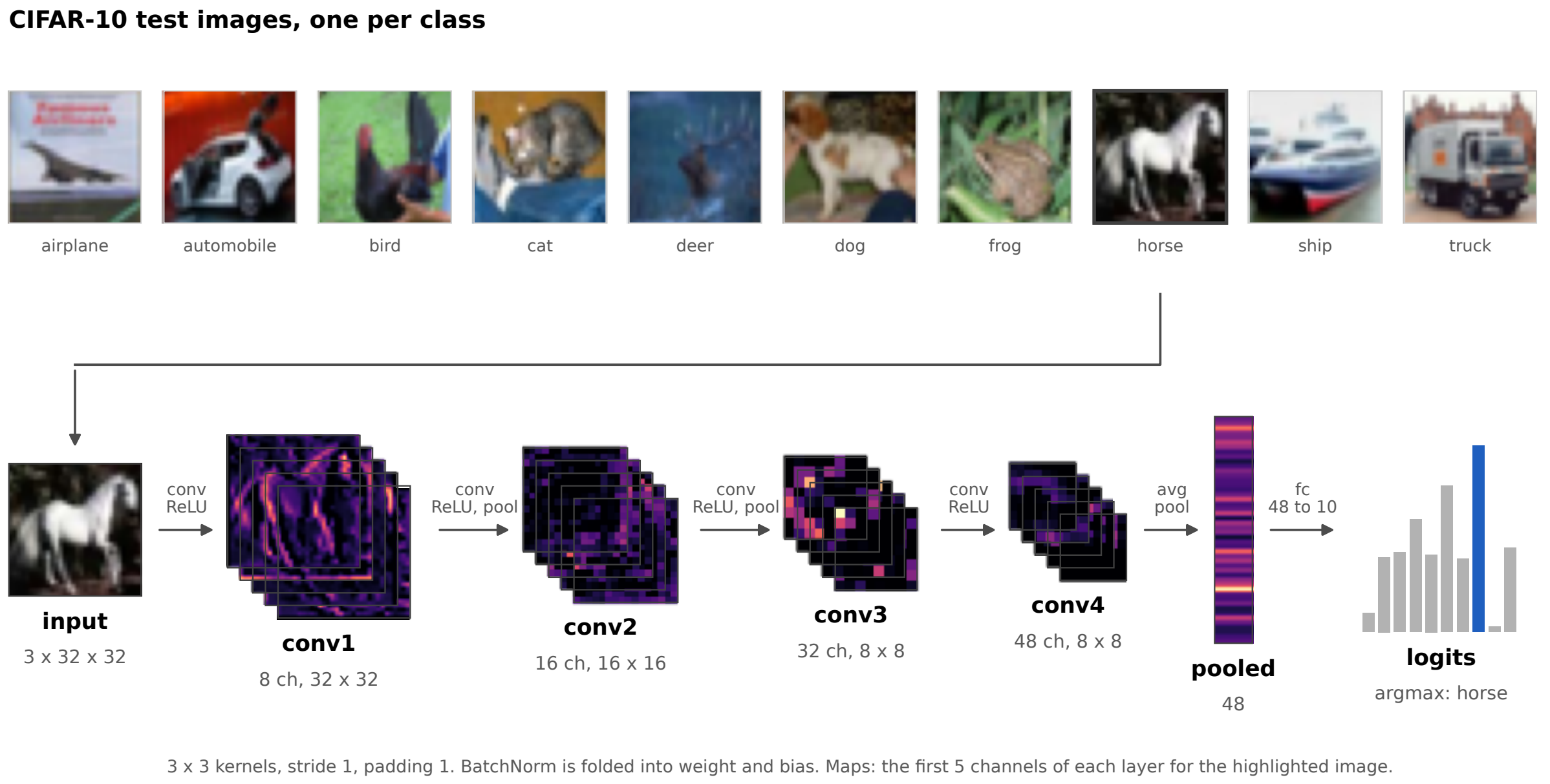

Figure 1: `SmallCNN`. One CIFAR-10 test image per class. The highlighted one is passed through the float network. Each stack shows the first five channels after ReLU (and pooling), the bar the 48 pooled features, and the bars at the right the ten logits.

where every operation must be carried out in chain-of-thought [20].

# 3 Method

## 3.1 Network, baseline and notation

`SmallCNN` has four conv(3×3)–BN–ReLU blocks with 8, 16, 32 and 48 channels, 2×2 max-pooling after blocks 2 and 3, global average pooling and a 48→10 linear layer (20,280 weights). It is trained on CIFAR-10 for 500 epochs (SGD, Nesterov, one-cycle, weight decay $5 \cdot 10^{-4}$, crop and flip) on a fixed 45k/5k train/val split and reaches 82.14% val and 81.42% test accuracy (Figure 1).

BatchNorm is folded into the convolutions. We call the weights that compute one output channel a *row*, which is one conv filter ($C_{\text{in}} \times 3 \times 3$ weights) or one row of the fc matrix (48 weights). Hence the network has 114 rows (8+16+32+48+10). The row is also the unit in which the LLMs read and write the network. The input lists each row with its scale and bias (Section A.1), edits address weights by layer and row, and in P3/P4 the LLM writes one line per row. Per-channel RTN gives each row $r$ with weights $w_{rj}$ its own scale:

$$s_r = \max_j |w_{rj}| \,/ q_{\max}, \quad q_{rj} = \text{round}\big(w_{rj}/s_r\big), \qquad (1)$$

where $q_{rj}$ is the quantized integer weight for $w_{rj}$, $s_r$ the per-row scale, and $s_r q_{rj}$ the dequantized weight used at inference. With $b$ bits, $q_{\max} = 2^{b-1} - 1$ ($q \in \{-7, ..., 7\}$ at int4). Biases and activations stay float (BN folding does not change per-channel RTN).

For a row with scale $s$ we write $u = w/s$ for a float weight in units of the quantization step, $q = \text{round}(u)$ for its integer weight and $e = q - u \in [-1/2, 1/2]$ for its rounding error. The *kernel sum* $E = \sum_k e_k$ adds the rounding errors over the nine positions $k$ of one 3×3 kernel (for fc layers, over a whole row).

## 3.2 Protocols

In every experiment, a frontier LLM (Section 3.3 lists them) receives a task description and `SmallCNN` as text, and returns a quantized version of `SmallCNN`. Appendices A.4, A.5, A.6 and A.7 give the task descriptions of P1–P4. The four protocols differ in what the LLM sees and what it writes.

**P1: whole network, one answer.** The LLM sees the float weights in units of the step, $u = w/s$ (two decimals), and the RTN integer weights $q$ for every row, plus RTN scales and biases (format in Section A.1). It is told the architecture, input normalization, and that the float network scores 82.14% on a held-out split while the quantized one scores 60.38% (int4). It answers with sparse edits that set one integer weight, replace a 3×3 kernel or an fc row, or set a scale or a bias. Unmentioned values keep their RTN value. Invalid lines are skipped, never repaired. Writing $u$ instead of $w$ saves one division per weight and exposes the rounding residual $u - q$ directly, which is a hint.

**P2: layer by layer.** Same input, but delivered in eleven chunks (conv1, conv2, conv3 in two, conv4 in six and fc, with at most 256 kernels each). The LLM answers each chunk in its own response, inside one conversation, so every chunk gets a fresh output budget while the context keeps everything so far.

**P3: from float weights.** The LLM gets only the float weights (three decimals), in 19 chunks of at most 1,152 weights, and writes the 4-bit network itself as one line `ROW <layer> <r> SCALE <s> Q <codes>` per output row, with free scale and integer weights in $[-7, 7]$. The prompt does not mention RTN or the accuracy gap. A missing row, a wrong number of integer weights or an out-of-range integer weight falls back to RTN for that row.

**P4: from float weights with BN statistics.** As P3, plus $\gamma$, $\beta$, running mean and variance for every conv channel. Published data-free methods use these statistics in place of data. DFQ [7] estimates the input means from them for bias correction, and ZeroQ [8] synthesizes calibration data from them. P4 tests whether the LLM uses them in a similar way. Biases stay fixed, so the statistics can only act through integer weights and scales.

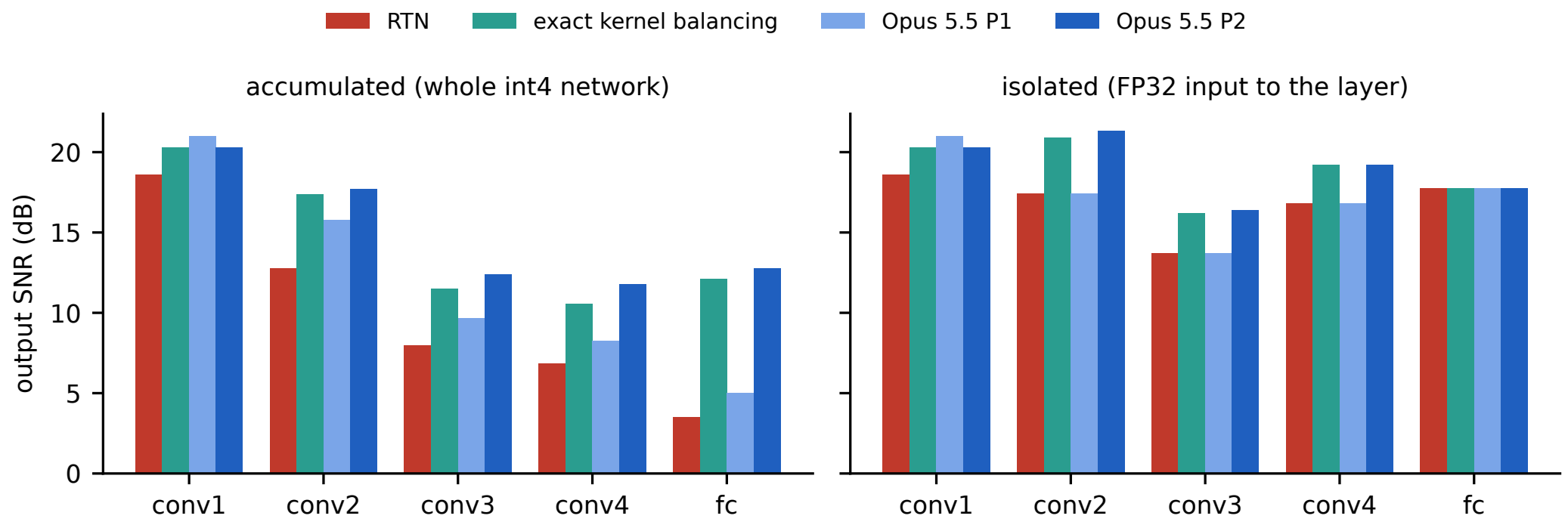

Figure 2: Output SNR per layer at int4. Left: accumulated. Right: isolated (float input to each layer). RTN, exact kernel balancing, Opus 5.5 P1 (conv1 only), Opus 5.5 P2.

### 3.3 LLMs and evaluation

*LLMs and tool restriction.* We use Claude Opus 5.5, Sonnet 5.5 and Haiku 5.5 in P1, P2 and P4, Opus 5.5 alone in P3, and Fable 5.2 only in P4 (API model IDs `claude-opus-5-5`, `claude-sonnet-5-5`, `claude-haiku-5-5`, `claude-fable-5-2`). P1–P3 ran as subagents that could call only the file-reading tool on their task files, which only delivers the input. We audited every transcript and found no other calls. P4 ran from the command line with all tools disabled (`claude -p --tools ""`), one chunk per message in one session, because API safeguards blocked the subagent route for Opus 5.5 and Sonnet 5.5.

*Evaluation.* We report test accuracy unless stated otherwise. For analysis only, we compute on the 5k validation images per-layer output SNR $10\log_{10}\left(\sum y_{\text{fp}}^2/\sum\left(y_q - y_{\text{fp}}\right)^2\right)$, *isolated* (float input to each layer) and *accumulated* (quantized network end to end), input means and patch covariances. The LLMs never see these.

*Reference corrections.* To judge the LLM edits, we also run two data-free corrections in Python. *Kernel balancing* re-rounds integer weights, cheapest first, until every kernel sum satisfies $|E| \leq 1/2$. This is the rule Opus 5.5 and Sonnet 5.5 describe in P1 (Section 4.2.1). We call this implementation, run on every kernel, *exact kernel balancing*. *Bias correction* [7] keeps the integer weights and cancels the mean output shift through the bias, using input means measured on validation images. In section Section 5.1.3 we provide more theoretical details about those two methods.

## 4 Results

### 4.1 What RTN does to the network

For our simple network, RTN is basically lossless at 8 bit (81.45%, 98.6% agreement with float) and drops by 21 percentage points (pp) to 60.39% at 4 bit (Table 1). Below 4 bit RTN gives 28.7% at 3 bit and chance at 2 bit. Below 4 bit no data-free correction comes near float accuracy, hence in the remainder of the paper **we study the 4-bit case, where corrections can recover most of the loss.**

Table 1: Test accuracy (%) of RTN and of the two corrections defined in Section 3.3, from 8 to 2 bit calculated using a Python script. The float network reaches 81.42%. Kernel balancing is applied to every conv kernel and fc row. Underlined: best value per column. Shaded: 4 bit, the setting we study in the rest of the paper. †Uses input means measured on validation images, so it is not data-free.

| Test % | 8 bit | 6 bit | 5 bit | 4 bit | 3 bit | 2 bit |
|---|---|---|---|---|---|---|
| RTN | 81.45 | 80.86 | 77.91 | 60.39 | 28.70 | 8.60 |
| + bias corr.† | 81.38 | 81.33 | 80.80 | 75.29 | 50.75 | 12.03 |
| kernel bal. | 81.41 | 81.55 | 80.10 | 77.80 | 62.47 | 13.26 |
| both† | 81.40 | 81.43 | 80.54 | 79.68 | 64.23 | 12.64 |

Table 2 compares kernel balancing (what the frontier LLMs propose in P1/P2) with published PTQ methods. Kernel balancing on the conv layers is basically SQuant's [9] kernel step and leads to 77.63%, and adding SQuant's channel step gives 79.06%. With 1,024 calibration images from the training split, AdaRound [4] and GPTQ [6] reach 80.7%, 0.7 pp below float. All methods keep the RTN scales, and all except bias correction change only the rounding.

Table 2: Test accuracy (%) at int4 of PTQ methods on our network (FP32: 81.42), our implementations, RTN per-row scales throughout. Calibration images come from the training split.

| Method | Data | int4 |
|---|---|---|
| RTN | none | 60.39 |
| kernel bal. = SQuant-K (conv) | none | 77.63 |
| kernel bal. (conv + fc rows) | none | 77.80 |
| SQuant-K+C [9] | none | 79.06 |
| bias corr. | val means | 75.29 |
| kernel bal. + bias corr. | val means | 79.68 |
| GPTQ [6] | 1k images | 80.66 |
| AdaRound [4] | 1k images | 80.71 |

The total accuracy loss is not caused by a single layer. The isolated SNR of all five int4 layers lies between 14 and 19 dB, while the accumulated SNR falls from 18.6 dB after conv1 to 3.5 dB at the logits (Figure 2). Every layer contributes a similar error, and the errors compound.

Table 4: Int4 LLM results, grouped by protocol with what the LLM receives (in) and returns (out). In P1 and P2 the LLM is told the float and RTN accuracies and asked to improve the RTN network. In P3 and P4 it is asked to quantize the float network and is told neither. Δ RTN: change in test accuracy relative to RTN, green if better, red if worse. Invalid rows (P3, P4) fall back to RTN.

| Protocol | LLM | Changed / valid | Test % | Δ RTN (pp) | What the LLM did |
|---|---|---|---|---|---|
| **Reference** computed in code | RTN | – | 60.39 | – | |
| | exact kernel bal. | 1,495 weights | 77.80 | +17.41 | code, all layers, no data (reference) |
| | AdaRound | 2,718 weights | 80.71 | +20.32 | code, 1,024 calibration images (reference) |
| **P1 whole network** in: float + RTN int4, held-out 82.1% → 60.4% out: edits, one answer | Opus 5.5 | 11 weights | 64.07 | +3.68 | kernel balancing, conv1 only |
| | Sonnet 5.5 | 13 weights | 64.37 | +3.98 | kernel balancing, conv1 only |
| | Haiku 5.5 | fc rows | 62.18 | +1.79 | balances fc row sums |
| **P2 layer by layer** in: as P1, in 11 chunks out: edits, one answer per chunk | Opus 5.5 | 1,364 weights | <u>77.95</u> | +17.56 | kernel balancing through all 2,200 kernels, 96% identical to exact kernel bal. |
| | Sonnet 5.5 | 0 | 60.39 | ±0.00 | declines with “no basis for changing any code” |
| | Haiku 5.5 | 74 weights, 11 scales | 59.90 | −0.49 | scales 11 rows by 0.9 on an untested weight-MSE argument |
| **P3 float only** in: float weights only out: full int4 network | Opus 5.5 | 111 / 114 rows | 60.58 | +0.19 | writes RTN, clips conv4 outliers (+1.5 pp), loses it on conv1 ties |
| **P4 float + BN** in: float weights + BN stats out: full int4 network | Fable 5.2 | 114 / 114 rows | <u>69.97</u> | +9.58 | row-wise clipping of isolated outliers in all layers, BN unused |
| | Opus 5.5 | 108 / 114 rows | 63.26 | +2.87 | clips conv4 and fc outliers, BN unused |
| | Sonnet 5.5 | 39 / 114 | 59.23 | −1.16 | RTN with wrong number of weights and shifted integer weights in valid rows |
| | Haiku 5.5 | 19 / 114 | 59.96 | −0.43 | wrong number of weights, otherwise copies RTN |

Table 3: RTN int4 layer statistics. Peak/RMS: ratio of largest to RMS float weight. Zeros: fraction of zero integer weights. SNR (dB): accumulated and isolated. Mean $|E|$: mean absolute rounding-error sum per 3×3 kernel (fc: per row), in steps.

| layer | weights | peak/RMS | zeros | SNR acc. | SNR iso. | mean $\|E\|$ |
|---|---|---|---|---|---|---|
| conv1 | 216 | 2.28 | .18 | 18.6 | 18.6 | 0.59 |
| conv2 | 1,152 | 3.03 | .22 | 12.8 | 17.4 | 0.71 |
| conv3 | 4,608 | 3.32 | .21 | 8.0 | 13.7 | 0.68 |
| conv4 | 13,824 | 3.23 | .19 | 6.8 | 16.8 | 0.73 |
| fc | 480 | 3.03 | .18 | 3.5 | 17.7 | 1.29 |

## 4.2 LLM results

Table 4 lists the int4 LLM results with the mechanism behind each result. We discuss them by protocol below.

### *4.2.1 P1: Editing the whole network*

In P1 (Table 4), Opus 5.5 and Sonnet 5.5 arrive at the same idea. Both argue that natural images are smooth, so a 3×3 kernel responds mainly to the sum of its weights. They compare float and integer sums per conv1 kernel and re-round the weight closest to a boundary where they differ by more than about half a step. This is kernel balancing and we analyze it in Section 5.1. They change 11 and 13 integer weights (0.05% of the weights, 8 in common) and gain 3.7–4.0 pp. Haiku 5.5 balances the fc row sums instead. None of the LLMs touches conv2–conv4. All state that the work does not fit into one answer and stop. The limit is the length of one answer, which must hold the reasoning and every edit. Their hand execution is accurate. Sonnet 5.5′s 13 edits match exact kernel balancing in 11 positions, and the edited conv1 layers reach the same mean $|E|$ as exact kernel balancing (0.27 vs. 0.59 for RTN). Applied to all conv layers, the same rule gives 77.63% (Table 2). P2 lifts the output limit for the LLM itself (Section 4.2.2).

### *4.2.2 P2: One answer per layer*

With one response per chunk, Opus 5.5 works through every kernel of every conv layer (13, 77, 300 and 974 edits) and reaches 77.95% test (78.62% val), which closes 17.6 of the 21 pp gap to float and matches exact kernel balancing within noise (Figure 4). Opus 5.5 did not edit fc, and exact kernel balancing on the conv layers alone gives 77.63%. The run was interrupted and its conv1 and conv2 edits were re-emitted after resumption (Section 6). 1,314 of its 1,364 edits coincide with exact kernel balancing in position and value, no integer weight moves by more than one step, and 118 of 2,200 kernels keep $|E| > 1/2$. Isolated SNRs match exact kernel balancing within 0.4 dB in every layer, and the accumulated SNR at the logits rises from 3.5 to 12.8 dB. Ahead of exact kernel balancing by 2.3 pp after conv2 and a few tenths at the end, Opus 5.5′s curve mainly shows that the choice among nearly equivalent flips moves accuracy by about 1 pp.

Opus 5.5′s kernel sums have the same distribution as those of exact kernel balancing (Figure 3). Sonnet 5.5 reads all chunks and edits nothing, citing missing activation statistics, although it found kernel balancing in P1. Haiku 5.5 shrinks eleven scales by 0.9 and loses 0.5 pp.

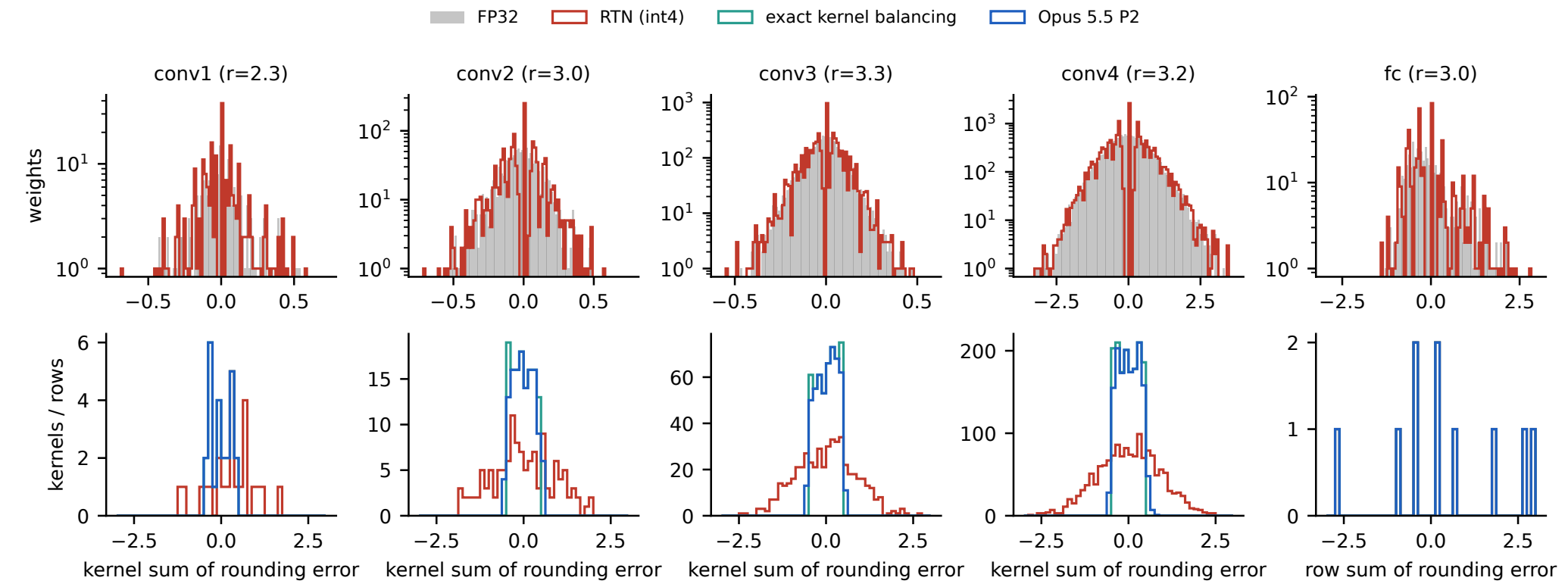

Figure 3: Top: float vs. dequantized int4 RTN weights (log count). Bottom: distribution of the per-kernel rounding-error sum $E$ (fc: per row) for RTN, exact kernel balancing and Opus 5.5 P2.

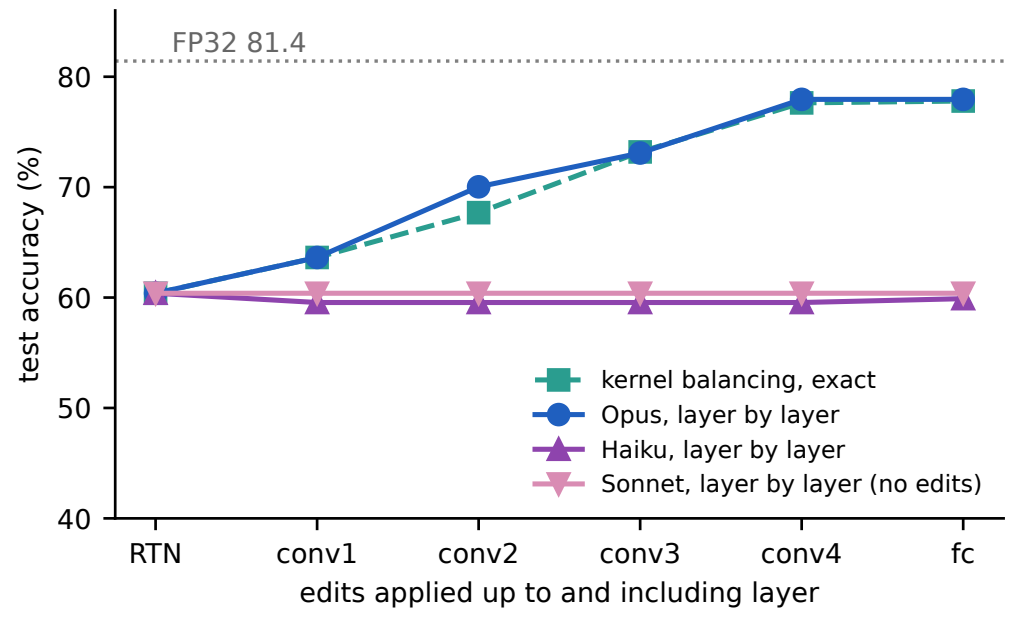

Figure 4: P2: cumulative test accuracy after editing up to a layer, Opus 5.5 vs. exact kernel balancing.

#### *4.2.3 P3/P4: Quantizing from float weights alone*

*Without the RTN result (P3)*, Opus 5.5 states the textbook recipe ($s = \max|w|/7$, "wastes none of the 15 levels") and executes it accurately. In each layer, 98.1–100% of integer weights equal round($w/s$) for its own scale, and 47 of the 48 exceptions lie within 0.02 of a tie. Its one departure is a lower scale (median 0.96× RTN) for conv4 rows with one or two outliers. It scores 60.58%, +0.2 pp over RTN. The conv4 clipping is worth +1.5 pp, while three conv1 integer weights rounded the other way at exact ties cost 1.1 pp. Three rows have the wrong length (143, 290, 287 integer weights) and fall back to RTN. Without the RTN integer weights and without being told that RTN fails, the LLM does not look for a problem.

*With BN statistics (P4)*, no LLM uses them (Table 4). Opus 5.5 and Fable 5.2 state that the statistics were only context. Sonnet 5.5 argues that they are already folded into the weights, which is true for the scales and misses that the running means describe activations. The statistics would have allowed one useful correction. They give the input means $\mu_c$ (Section 5.1.4), and with them each row can be rounded so that its mean shift in Equation 3 cancels. None of the LLMs does this.

*Opus* (63.26%) repeats its P3 strategy with more clipping, with median scale ratios of 0.96 (conv4) and 0.97 (fc). Per-layer swaps into RTN give +2.0 pp (conv4) and +1.5 pp (fc). Re-rounding the float weights exactly at its scales gives 63.07%, so the gain is in the scales. An MSE-optimal per-row scale search on conv4 and fc gives 64.39%, which Opus 5.5 approximates by eye.

*Fable 5.2* (69.97%, all 114 rows valid) is the best LLM result without the RTN integer weights. Its summary describes "per-row round-to-nearest with a mildly clipped scale" and adds "when one weight was an isolated outlier, I picked a smaller scale and let that weight clip to ±7". The data confirm this. In conv2, conv4 and fc the scale ratio correlates with the ratio of largest to second-largest weight at −0.87, −0.85 and −0.90, exact rounding at its scales gives 69.94%, and kernel sums are not reduced. Section 5.2 analyzes this result.

*Sonnet and Haiku* choose a workable method and fail at execution. Both write plain max/7 RTN, but most rows have the wrong number of integer weights or are cut off (19–39 of 114 valid). Some valid Sonnet 5.5 rows contain shifted integer weights, and its valid conv4 rows have an RMS error of 1.7 steps against 0.29 for RTN. Haiku 5.5 often does not return the correct number of integer weights, and otherwise more or less copies the input. Its valid rows equal RTN in 98–100% of the integer weights per layer.

### 4.3 Inference cost

For comparison, kernel balancing in Python runs in milliseconds on a CPU, and AdaRound needs a few minutes on one GPU. Table 5 lists the approximate price of the LLM runs. We used a Claude Pro subscription, so those are estimates based on the number of tokens. All API calls of the experiments together produced 3.6M output tokens and would probably cost about $81 at list API prices. The P4 runs account for two thirds, because each LLM writes every integer weight of the network. Output tokens dominate wherever an LLM computes by hand, because summing nine residuals per kernel and writing the result costs reasoning tokens for every kernel.

The two best runs differ in price per gain. Fable 5.2 in P4 wrote 272k output tokens over 21 responses, about 90% of them hidden reasoning, and read 4.2M input tokens, almost all from the prompt cache. At $24.7 ($10.9 output, $13.2 cache writes) it gains 9.6 pp. Opus 5.5 in P2 wrote 447k output tokens over 15 responses with a context of up to 677k tokens. At about $13 it gains 17.6 pp. The P1 run of Opus 5.5 wrote 134k output tokens for 3.7 pp.

Table 5: Output tokens and cost at list API prices per 1M tokens (input/output): Opus 5.5 \$4/\$20, Sonnet 5.5 \$2/\$10, Haiku 5.5 \$0.50/\$2.50 (context above 100k), Fable 5.2 \$8/\$40. Cache writes cost 1.25× and cache reads 0.05× the input price. P3 and P4 from the logged usage of all API calls. [a]Estimated from logged output tokens and maximum context, assuming linear context growth.

| Protocol | Output tokens | Cost (USD) |
|---|---|---|
| P1 whole network | 178k | 5[a] |
| P2 layer by layer | 525k | 15[a] |
| P3 float only | 235k | 7 |
| P4 float + BN | 2,642k | 53 |
| Total | 3,580k | 81 |

*Input encoding.* The int4 payload is 161k characters. Kernel balancing needs only the integer weight and a coarse residual. Running it on the residual quantized to three levels per step gives 78.68%, and on its sign alone 66.45%. A packed encoding (one letter for $q$, one for round$(10(u-q))$) needs a quarter of the characters. We kept the neutral format, since a compressed one would hint at the method.

## 5 What the LLMs did and why it works

The LLMs that gain over RTN use one of two mechanisms. In P1/P2 they keep the RTN scales and change the rounding of single weights (kernel balancing). In P3/P4 they keep plain rounding and change the scales (outlier clipping). This section explains why each works and why the LLMs probably chose them.

### 5.1 Kernel balancing

Two observations call for an explanation. First, the per-weight error of int4 RTN is small (about 13% relative RMS, Table 3), yet accuracy drops by 21 pp. Second, kernel balancing makes individual weights *less* accurate, so per-weight error cannot be what it optimizes. A first-order model of the rounding error explains both and makes predictions that we test on validation data. A conv output is $y_o = \sum_c \sum_k W_{ock} x_{ck}$ over input channels $c$ and kernel positions $k$. With the weight error $\Delta W = sq - w = se$, the output error is $\Delta y_o = \sum_c \sum_k \Delta W_{ock} x_{ck}$.

#### *5.1.1 Why the kernel sum matters*

The output error of one kernel is $\Delta y = s \sum_k e_k x_k$, so each rounding error is weighted by the input it multiplies. Consider a constant input patch, where the nine inputs under the kernel (pixels or feature-map activations) are equal, $x_k = x$. Then $x$ factors out, and $\Delta y = sx \sum_k e_k = sxE$ with $E = \sum_k e_k$ the rounding-error sum of the kernel. Which weight carries which error no longer matters, only the total. Figure 5(a) shows two error patterns. In kernel A the individual errors are large ($\|e\|^2 = 0.50$) but cancel, $E = 0$, so a constant patch produces no output error. In kernel B every error is small ($+0.1$, $\|e\|^2 = 0.09$) but all have the same sign, so they add up to $E = 0.9$ and the output is off by $0.9sx$. Per-weight error and output error rank the two kernels in opposite order. RTN minimizes every $|e_k|$ separately and never checks whether the errors cancel.

Of course real patches are not always close to constant (e.g. at edges). Writing each input as the patch mean plus a deviation, $x_k = \bar{x} + d_k$ with $\sum_k d_k = 0$, splits the output error exactly into

$$\Delta y = s\bar{x}E + s\sum_k e_k d_k. \tag{2}$$

The first term depends only on the kernel sum. The second comes from edges and texture inside the patch and depends on the individual errors. Kernel balancing will not reduce it and might even increase it slightly. Natural images and feature maps are mostly smooth, so the first term dominates, except at edges. Section 5.1.4 measures both terms. Equation 4 below quantifies this balance through the input correlation $\rho$. That a kernel responds mainly to the sum of its weights is also the argument Opus 5.5 and Sonnet 5.5 give for their edits in P1 (Section 4.2.1).

#### *5.1.2 Mean shift and fluctuating error*

The constant-patch argument has two consequences on real data. Averaged over positions and images, each input channel $c$ has a mean input $\mu_c = \mathbb{E}[x_{ck}]$ (about 0 for the normalized image, positive after ReLU), which takes the place of $x_c$ and gives an expected output shift

$$\mathbb{E}[\Delta y_o] = \sum_c \left( \sum_{k\in 3\times 3} \Delta W_{ock} \right) \mu_c = s_o \sum_c E_{oc}\mu_c, \tag{3}$$

a bias that is the same at every position and in every image, so it does not average out. What remains, $\Delta y_o - \mathbb{E}[\Delta y_o]$, is a fluctuating part. It changes with the content of each patch and averages to zero, like noise. A bias correction can cancel the mean but not this part. Its size depends on how the inputs vary together. Assume the nine inputs of a kernel have variance $\sigma^2$ and common correlation $\rho$. Then the variance of the fluctuating part, in units of $s^2$, is

$$e^\top \Sigma e = \sigma^2 \left[ (1-\rho)\|e\|_2^2 + \rho E^2 \right], \tag{4}$$

which interpolates between the per-weight error $\|e\|_2^2$ (uncorrelated inputs, $\rho = 0$) and the kernel sum $E^2$ (constant patches, $\rho = 1$). Section A.2 relates $\sigma^2$ and $\rho$ to the two terms of Equation 2. RTN minimizes every $|e_k|$ but leaves $E$ unconstrained. With nine i.i.d. uniform errors, std$(E) \approx 0.87$ steps.

(a) Two error patterns $e$ on a constant patch

kernel A

| | | |
|---|---|---|
| +.4 | −.4 | 0 |
| +.3 | −.3 | 0 |
| 0 | 0 | 0 |

$\|e\|^2 = 0.50$
$E = 0$, $\Delta y = 0$

kernel B

| | | |
|---|---|---|
| +.1 | +.1 | +.1 |
| +.1 | +.1 | +.1 |
| +.1 | +.1 | +.1 |

$\|e\|^2 = 0.09$
$E = 0.9$, $\Delta y = 0.9sx$

(b) Kernel balancing on one kernel

float $u = w/s$

| | | |
|---|---|---|
| 2.58 | −0.62 | 1.55 |
| 0.57 | 3.40 | −1.62 |
| −2.44 | 0.61 | 1.62 |

RTN $q$ (error $e$)

| | | |
|---|---|---|
| 3<br>+.42 | −1<br>−.38 | 2<br>+.45 |
| 1<br>+.43 | 3<br>−.40 | −2<br>−.38 |
| −2<br>+.44 | 1<br>+.39 | 2<br>+.38 |

$\|e\|^2 = 1.50$
$E = +1.35$

balanced (error $e$)

| | | |
|---|---|---|
| 3<br>+.42 | −1<br>−.38 | 1<br>−.55 |
| 1<br>+.43 | 3<br>−.40 | −2<br>−.38 |
| −2<br>+.44 | 1<br>+.39 | 2<br>+.38 |

$\|e\|^2 = 1.60$
$E = +0.35$

Figure 5: (a) On a constant input patch ($x_k = x$) the output error is $sxE$. Kernel A has the larger per-weight error but its errors cancel. Kernel B has small errors of one sign that add up. (b) RTN errors that happen to be mostly positive sum to $E = +1.35$ steps. Moving one weight down by one step lowers $E$ by 1. Balancing picks the weight with the largest positive error, i.e. the one closest to its rounding boundary (1.55, error +.45), and rounds it down from 2 to 1 instead of up. The squared error grows by $1 - 2 \cdot 0.45 = 0.10$, but on a constant patch the output error falls from $1.35sx$ to $0.35sx$.

#### 5.1.3 Two corrections

*Bias correction* [7] keeps the integer weights and changes the bias. It adds $\delta b_o = -\mathbb{E}[\Delta y_o]$ from Equation 3, which removes the mean shift. It needs the input means $\mu_c$, from data or from the BN parameters, and leaves the fluctuating part of the error unchanged.

*Kernel balancing* (Figure 5(b)) keeps the bias and changes integer weights until every kernel has $|E| \leq 1/2$. Rounding a weight to its other neighboring integer changes its error, and therefore $E$, by exactly 1. A kernel with $E = +1.35$, for example, reaches $E = +0.35$ when one weight that was rounded up is rounded down instead. The weights to change are chosen by cost. Changing a weight with error $e_k$ raises its squared error by $1 - 2|e_k|$. The cheapest weights are those with $|e_k|$ close to $1/2$, whose float value lay almost halfway between two integers. Balancing therefore changes the weights whose error has the same sign as $E$ and the largest magnitude. It needs no data and changes few integer weights. Because $E$ appears in both Equation 3 and Equation 4, a smaller $E$ lowers the mean shift for any input means and lowers the correlated part of the error whenever $\rho > 0$. For fc layers we apply the same rule to the sum over the whole row.

Kernel balancing is the kernel-wise step of SQuant [9], and our implementation produces the same integer weights (*exact kernel balancing* in the tables). SQuant motivates this step with a Hessian approximation that is constant within a kernel, which corresponds to $\rho > 0$ in Equation 4. Equation 3 gives a second reason for it, because the same sum sets the mean shift.

#### 5.1.4 Validation on data

We test the error model of Section 5.1 on validation data. The mean shift predicted by Equation 3 from measured input means matches the measured pre-activation shift with correlation ≥ 0.994 in every layer. At int4 it is 7–11% of the output standard deviation. For the unpooled inputs of conv2 and fc the input means can also be predicted from BN parameters alone (a clipped Gaussian with mean $\beta$ and std $|\gamma|$, correlation 0.95 and 0.92), as in DFQ [7]. We did not test the max-pooled inputs of conv3 and conv4. Since biases are fixed in P4, the LLMs could use these means only through the rounding (Section 4.2.3).

Table 6 splits the expected squared layer error, using measured patch covariances. The mean term is about 40% of the RTN error. Bias correction removes it and nothing else. Kernel balancing removes about 90% of it and makes the off-diagonal term slightly more negative, so both corrections reach nearly the same total error (.014 and .013). A relative error of only 0.024 thus costs 21 pp, because a small systematic error matters more than a larger random one.

Table 6: Expected squared layer error at int4 relative to the float pre-activation variance (mean over layers, float inputs), split into mean, diagonal and off-diagonal parts.

| int4 | mean | diag | off | total |
|---|---|---|---|---|
| RTN | .010 | .018 | −.004 | .024 |
| + bias corr. | .000 | .018 | −.004 | .014 |
| kernel bal. | .001 | .019 | −.007 | .013 |
| both | .000 | .019 | −.007 | .012 |

The split of Equation 2 shows where balancing helps (Figure 6). Under RTN the patch-mean term, which holds the mean shift of Table 6 plus the fluctuation of the patch mean, carries 53–76% of the squared output error in conv1–conv4. Balancing cuts this term to 9–30% of its RTN value, while the detail term grows by 21–46%. On the smoothest tenth of patches it removes 68–87% of the error in conv1–conv3, while on the sharpest edges in conv1 it increases the error by half, because there the individual errors matter.

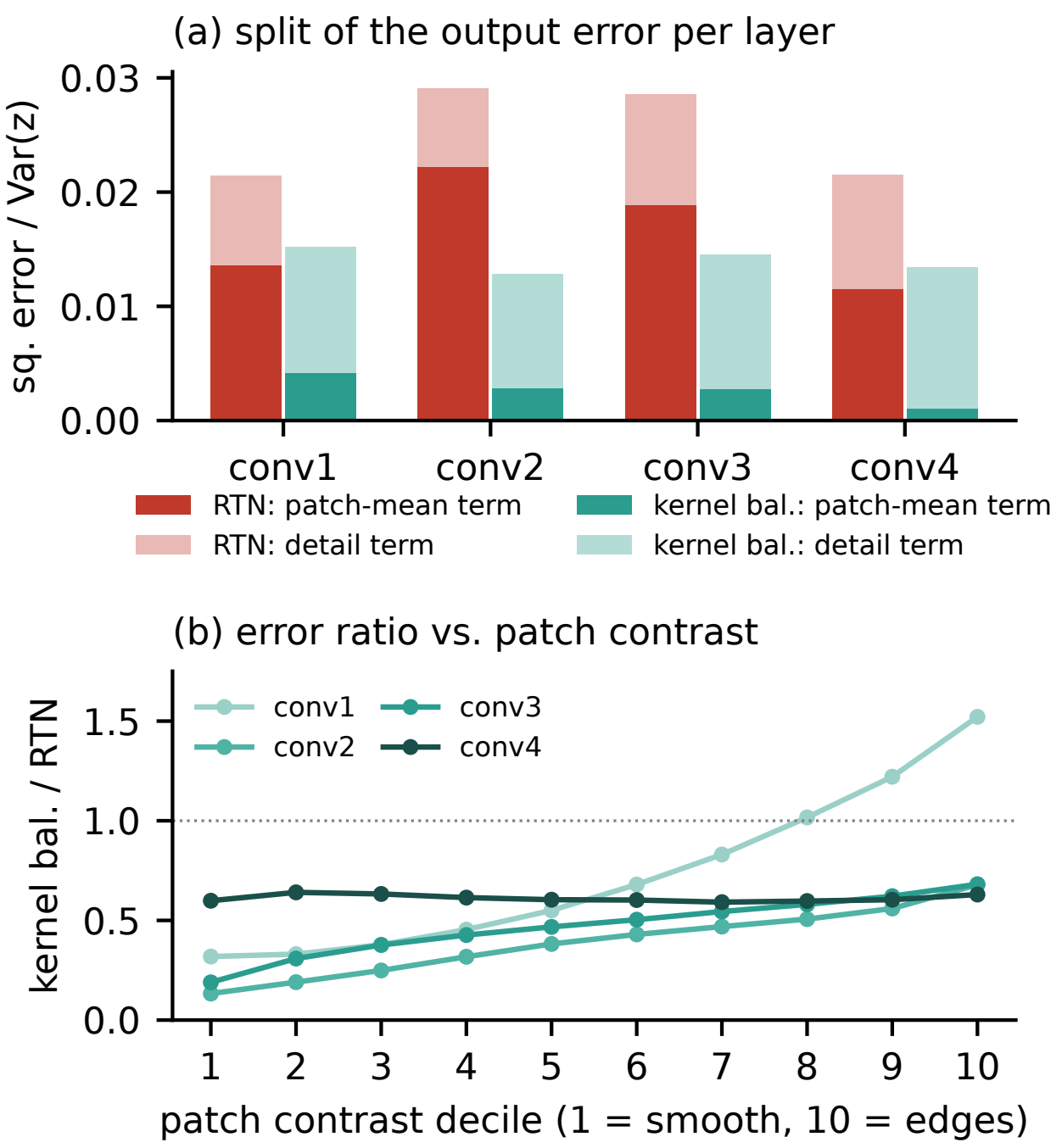

Figure 6: Exact split of the int4 output error (Equation 2) on validation patches. (a) Mean-square error relative to the float pre-activation variance, split into the patch-mean term (solid) and the detail term (light), for RTN and kernel balancing. (b) Error of kernel balancing relative to RTN, with patches sorted into deciles by their contrast, from smooth (1) to edges (10).

### 5.2 Outlier clipping

Fable 5.2's P4 result (Section 4.2.3) comes from the scales alone. Exact rounding at its scales gives 69.94%, and its kernel sums are not reduced. It clips less than MSE search (median ratios in conv2–fc 0.95–0.98 vs. 0.80–0.92, both about 0.99 in conv1), yet MSE search on all layers gives only 64.6%. A possible explanation, which we did not test, is that clipping one isolated outlier gives the bulk of a heavy-tailed row a finer grid at small cost, while clipping a group of large weights, which MSE does, puts the error on the weights that matter most.

The gain is uneven across layers, with +8.8 pp in conv2, +2.8 in conv4, +0.3 in fc, −2.7 in conv1 and −2.0 in conv3 (72.1% without its conv1 rows). Conv2 is unusually sensitive. Fable 5.2's conv2 rows alone give 69.2%. Random per-row scale factors in $[0.88, 1]$ give 66.1% on average (61.4–69.0%, 12 draws), and Fable 5.2's ratios shuffled across rows 64.3% (60.1–68.0%, 12 shuffles). Its assignment beats all 24 controls, but the best random draw comes within 0.3 pp, and with one run we cannot quantify how much luck it holds. Random factors already recover 5.7 of the 8.8 pp, so most of the conv2 gain comes from re-rounding at a slightly smaller scale, not from the choice of rows.

Opus 5.5 does the same more weakly in P4, in conv4 and fc only (63.26%, Section 4.2.3). Outlier clipping is a standard refinement of RTN, applied by MSE search in most toolchains [3]. Fable 5.2′s choice of rows adds about 3 pp over random factors in conv2 and costs accuracy in conv1 and conv3.

### 5.3 Why did the LLMs use these methods?

*What the input allows.* In P1/P2 the LLM may change integer weights, scales and biases, but it sees no data and no BN statistics. Bias correction needs the input means $\mu_c$ of Equation 3, so it would require guessing them. Scale changes need an error criterion that the LLM would have to evaluate by hand for every candidate scale. The kernel sum is the one quantity that predicts the output error without knowing the inputs, under the single assumption that inputs are smooth, and the P1/P2 format shows every residual $u - q$ next to its integer weight. Kernel balancing is therefore close to the cheapest correction that the input supports. Haiku 5.5 states the mean-shift argument of Equation 3 directly for the fc layer. Its inputs "are non-negative after ReLU and global average pooling, so the sum of rounding errors in each row acts as a systematic shift on the logits".

*Recall or derivation.* No LLM names SQuant, but Opus 5.5 places its rule in the literature, writing "The idea follows the bias-correction and adaptive-rounding view of post-training quantization." Our runs therefore cannot tell whether the rule is derived from the weights or recalled and then recognized as applicable. A prompt that describes the setting without any weights would separate the two.

*Known and declined.* In P3/P4 the LLMs list the corrections and decline them. Opus 5.5 writes "Biases are fixed and there is no data, so I couldn't do bias correction, activation-aware rounding, or cross-layer equalization." Sonnet 5.5 writes "I didn't use bias correction, cross-layer equalization, or any rounding that cancels error across a row." Sonnet 5.5 thus names error-cancelling rounding in P4 and does not apply it. Both treat plain rounding as an adequate answer when nothing says that it fails. Section 7 discusses which differences between the prompts may cause this.

## 6 Limitations

*Statistics.* Every LLM number comes from one run of a particular LLM version. The binomial standard error is about 0.5 pp at 60%, and single flips in conv1 or conv2 move accuracy by about 1 pp. Differences below about 1 pp, such as the P3 result and the Haiku 5.5 results, are therefore within noise. We ran no paired test and no control with random near-tie flips for P1.

*Scope.* All four LLMs are Claude models. The network is small, and other architectures will have different rounding statistics. The input format of P1/P2 exposes the rounding residual, which hints at rounding-aware corrections. P4 ran in a different harness, and we have no P3 runs for Sonnet 5.5, Haiku 5.5 and Fable 5.2, so the effect of the BN statistics on their behavior is not isolated.

*Execution.* The tool restriction in P1–P3 rests on instructions and audited transcripts, and hidden reasoning is not visible. Some runs were interrupted by usage limits and resumed, and in two runs the LLM's first output was only in hidden reasoning and had to be re-emitted. The Opus 5.5 P2 run stopped after chunk 10. Its conv1 and conv2 edits had been written only in hidden reasoning and were re-emitted as text after resumption (without them it scores 69.58%), and it did not redo the fc chunk.

*Proprietary APIs.* Midway through the experiments the API safety classifiers began to reject our runs. Requests of the same kind that had run before were stopped at the first message, for P4 with Opus 5.5 and Sonnet 5.5 as subagents and for a P3 rerun. We had to change the harness and try several prompt variants until the runs were accepted again, which is why P4 ran from the command line (Section 3.3). We do not know why the classifier fired. Such changes are invisible to the user, can stop an experiment at any point and make it hard to reproduce. This is a general problem of research run against proprietary APIs.

*Next steps.* Natural next steps are repeated runs with paired tests, a random-flip control for P1, a bias-correction reference from BN statistics alone, prompts that state the RTN gap in P3/P4, a P1 prompt with raw float weights, a prompt that describes the task without weights (to separate recall from derivation), a small evaluation budget, and a compressed delivery format.

## 7 Conclusion

**Without tools, an LLM can repair part of a quantized CNN by only reading its weights**. At 4 bit two LLMs use kernel balancing (the kernel step of SQuant), a data-free rule that removes most of the mean shift our measurements show in every layer. Delivered one layer per answer, Opus 5.5 works through all 2,200 kernels by hand and reaches 77.95%, 3.5 pp below float. Tasked to output only a single answer the LLMs found the method but could apply it only to conv1, because the arithmetic for the other layers does not fit. When the LLM must quantize float weights itself, the gains come from row-wise outlier clipping, and Fable 5.2 scores above an MSE scale search (70.0% vs. 64.6%), partly by luck (Section 5.2). Rounding that cancels the mean shift, for which the BatchNorm statistics given in P4 suffice at least in conv2 and fc, appears in no run. Calibrated methods (AdaRound, GPTQ) reach 80.7% on the same network, so the best LLM result is 2.8 pp short of methods that use calibration data.

*Why P3 and P4 did not perform rounding correction.* Kernel balancing appears only when the LLM sees the RTN network and is told that it fails, and even then not in every run (Sonnet 5.5 and Haiku 5.5 in P2). In P3 and P4 every LLM writes plain rounding, at most with clipped scales.

- *Stated defect.* P1/P2 report that rounding lowers accuracy on a held-out split from 82.1% to 60.4%, which turns the task into the search for a defect. P3/P4 report only the float accuracy, and the LLMs treat plain rounding as the expected answer (Fable 5.2 writes "I expect it to behave like a reasonable 4-bit per-channel post-training quantization").
- *Visible residual.* P1/P2 list every weight in units of the step, $u = w/s$, next to its integer $q$, so the rounding error $u - q$ can be read off and summed per kernel. In P3/P4 each residual first requires a division by a scale that the LLM chooses itself.

- *Output volume.* P1/P2 accept sparse edits. P3/P4 require all 20k integer weights, so most of the output goes into writing integers.
- *Reference rounding.* The RTN integers show what plain rounding produces and invite a comparison of integer and float sums, which is how Opus 5.5 and Sonnet 5.5 describe their P1 idea.

Fable 5.2 came close in P4, writing that it was "sometimes alternating direction within a row so the row sum would not drift one way". A P3 prompt that states the RTN gap would test the first factor, and a P1 prompt with raw float weights the second.

Code, prompts, replies and result files are available at `github.com/urbste/weight-space-programming-ptq`.

*Use of AI tools.* The idea of the study, the four protocols and the experiments are the author's. AI tools were used to write the protocol prompts and the evaluation code, to carry out the experiments and helped to write the paper. The author read and corrected the paper and takes responsibility for its content.

## Appendix

### A.1 Input format and protocol

P1/P2 input (excerpt, int4):

```
conv1 r0 scale 0.0635392 bias 0.3771
conv1 r0 u  c0: 3.83 -1.47 0.37 -4.49 2.26 -0.49
0.82 ...
conv1 r0 q  c0: 4 -1 0 -4 2 0 1 -1 -1 | c1: 0 0 0
-7 ...
```

Edit language: `Q <layer> <row> <in_ch> <pos> <code>`, `K <layer> <row> <in_ch> <9 codes>`, `R fc <row> <48 codes>`, `S <layer> <row> <scale>`, `B <layer> <row> <bias>`. P2 instruction: "In EACH response read exactly one chunk file, write the edit lines for its rows between `BEGIN CHUNK k` and `END CHUNK k`, then read the next chunk." P3/P4 answer: `ROW <layer> <r> SCALE <s> Q <codes>`. Tool rule (P1–P3): "The only tool you may call is the file-reading tool, and only on this single file. […] no terminal, no code execution, no Python, no calculator, no file writing, no web, no subagents."

### A.2 Input variance and the two error terms

In Equation 4, $\sigma^2$ is the variance of a single input across patches, measured from the channel mean $\mu_c$, not the within-patch deviation $d_k$ of Equation 2. The two are linked by $\rho$. The patch mean has variance $\mathrm{Var}(\bar{x}) = \sigma^2(1 + 8\rho)/9$ and each deviation $\mathbb{E}[d_k^2] = \sigma^2 \cdot 8(1-\rho)/9$. Hence $\rho$ sets how much of the input variation reaches the kernel-sum term of Equation 2 and how much reaches the detail term.

### A.3 Token usage

Table 7: Responses, output tokens and maximum context per run. Not listed: the P4 runs of Opus 5.5, Sonnet 5.5 and Haiku 5.5, which together wrote 2,370k output tokens.

| Run (int4) | Resp. | Output | Max. ctx |
|---|---|---|---|
| Opus 5.5 P1 | 9 | 134k | 239k |
| Sonnet 5.5 P1 | 6 | 11k | 239k |
| Haiku 5.5 P1 | 15 | 33k | 238k |
| Opus 5.5 P2 | 15 | 447k | 677k |
| Sonnet 5.5 P2 | 13 | 2k | 238k |
| Haiku 5.5 P2 | 13 | 74k | 260k |
| Opus 5.5 P3 | 22 | 235k | – |
| Fable 5.2 P4 | 21 | 272k | 409k |

### A.4 Task description (P1, int4)

Figure 7 reproduces the P1 task description at int4 verbatim, followed by the start of the network dump. The P2 description is the same with the chunk instructions added. P3 and P4 replace the RTN part with the float weights and the `ROW` answer format. All prompts are in the project repository.

```
You are given the complete trained float32 weights of a small convolutional network for CIFAR-10
(10 classes: airplane, automobile, bird, cat, deer, dog, frog, horse, ship, truck), together with its
standard 4-bit post-training-quantized version. Your task: edit the 4-bit network so that it
classifies as accurately as possible.

RULES
- No tools: no code execution, no calculator, no files, no internet. Do every computation in your own
  reasoning and write the numbers directly.
- You cannot run the network. There is no data and there are no BatchNorm statistics: the weights below
  are all you get.
- You may only change the quantized network's parameters: integer codes, per-row scales and per-row biases.
  The result must remain a valid 4-bit network: integer codes in [-7, 7], one positive float
  scale per output row, one float bias per output row. The architecture stays as it is.

NETWORK
Input: 3x32x32 RGB image, normalized per channel with mean (0.4914, 0.4822, 0.4465), std (0.2470, 0.2435, 0.2616).
  conv1: Conv2d 3->8,   3x3, padding 1, bias, then ReLU
  conv2: Conv2d 8->16,  3x3, padding 1, bias, then ReLU, then MaxPool 2x2
  conv3: Conv2d 16->32, 3x3, padding 1, bias, then ReLU, then MaxPool 2x2
  conv4: Conv2d 32->48, 3x3, padding 1, bias, then ReLU, then global average pooling
  fc:    Linear 48->10 (logits; prediction = argmax)
The network was trained with BatchNorm after every conv; BatchNorm has been folded into the conv weights and
biases, so the network above is complete as written.

4-BIT FORMAT AND STARTING POINT
Every layer is stored as integer codes q in [-7, 7], one positive float scale per output row and one
float bias per row. The effective weight is w = q * scale[row].
The given 4-bit network is standard per-channel symmetric round-to-nearest PTQ:
scale[row] = max|w_float[row]| / 7, q = round(w_float / scale[row]), biases copied unchanged.
On a held-out split the float32 network scores 82.14% and this 4-bit network scores 60.38%.

DUMP FORMAT (below)
For every layer and output row r:
  "<layer> r<r> scale <s> bias <b>"                                  the RTN scale and the float bias
  "<layer> r<r> u ...", one or more lines                             FLOAT32 weights in units of the row's
                                                                       step: w_float = u * scale
  "<layer> r<r> q ...", one or more lines                             the RTN INTEGER codes (q = round(u))
conv layers: values are grouped per input channel; each group is "c<i>:" followed by the 3x3 kernel in
row-major order (position p = 3*kh + kw, 0..8). Long rows are split over several lines (c0-c7, c8-c15, ...).
fc: values are input features 0..47 in order, split over lines of 16 ("i0:" marks the first index on the line).

EDIT LANGUAGE (your answer)
  Q <layer> <row> <in_ch> <pos> <code>     set one conv code (pos 0..8)
  Q fc <row> <in> <code>                   set one fc code
  K <layer> <row> <in_ch> <9 codes>        replace a whole 3x3 kernel (conv only)
  R fc <row> <48 codes>                    replace a whole fc row
  S <layer> <row> <scale>                  set a row's scale (positive float)
  B <layer> <row> <bias>                   set a row's bias (float)
Rows/channels/positions are 0-based. Anything you do not edit stays as in the RTN network. If a parameter
is edited twice, the later line wins. There is no fixed limit on the number of edits, but your whole answer
must fit in a single response, so spend your effort on the edits you expect to matter most.

ANSWER FORMAT
A short explanation (a few sentences) of what you changed and why, then
BEGIN
<edit lines, one per line>
END
TOOLS USED: <every tool call you made>


## conv1 weight shape [8, 3, 3, 3]
conv1 r0 scale 0.0635392 bias 0.3771
conv1 r0 u  c0: 3.83 -1.47 0.37 -4.49 2.26 -0.49 0.82 -0.91 -0.55 | c1: 0.14 0.49 -0.50 -7.00 6.61 1.87 -1.1 ...
conv1 r0 q  c0: 4 -1 0 -4 2 0 1 -1 -1 | c1: 0 0 0 -7 7 2 -1 0 1 | c2: -1 -1 -1 -6 5 1 -1 2 1
[... 7 more conv1 rows, then conv2, conv3, conv4 and fc in the same format; 803 lines in total]
```

Figure 7: P1 task description at int4 as sent to the LLM. The dump lines of conv1 row 0 are cut here; the LLM receives them in full.

### A.5 Task description (P2)

P2 uses the P1 task description up to the edit language and replaces the answer format with chunked delivery (Figure 8). Each chunk file holds the same dump format as P1.

```
[... lines up to EDIT LANGUAGE identical to P1 (Figure 9), except "DUMP FORMAT (below)"
 reads "DUMP FORMAT (in the chunk files)" ...]

EDIT LANGUAGE (your answer)
  Q <layer> <row> <in_ch> <pos> <code>     set one conv code (pos 0..8)
  Q fc <row> <in> <code>                   set one fc code
  K <layer> <row> <in_ch> <9 codes>        replace a whole 3x3 kernel (conv only)
  R fc <row> <48 codes>                    replace a whole fc row
  S <layer> <row> <scale>                  set a row's scale (positive float)
  B <layer> <row> <bias>                   set a row's bias (float)
Rows/channels/positions are 0-based. Anything you do not edit stays as in the RTN network. If a parameter
is edited twice, the later line wins. There is no fixed limit on the number of edits, but your whole answer
is spread over one response per chunk (see below).

ANSWER FORMAT
Per chunk, as described under DELIVERY IN CHUNKS.

DELIVERY IN CHUNKS (this experiment)
The dump is NOT in this file. It is delivered in 11 chunk files, read one after another:
  chunk01.txt: conv1 rows 0-7
  chunk02.txt: conv2 rows 0-15
  chunk03.txt: conv3 rows 0-15
  chunk04.txt: conv3 rows 16-31
  chunk05.txt: conv4 rows 0-7
  chunk06.txt: conv4 rows 8-15
  chunk07.txt: conv4 rows 16-23
  chunk08.txt: conv4 rows 24-31
  chunk09.txt: conv4 rows 32-39
  chunk10.txt: conv4 rows 40-47
  chunk11.txt: fc rows 0-9
Work strictly chunk by chunk. In EACH response:
  1. read exactly one chunk file (the next one in order),
  2. think about that chunk only (you may use everything you have seen so far),
  3. write the edit lines for the rows of that chunk as plain text, between a line "BEGIN CHUNK <k>" and a
     line "END CHUNK <k>" (write "BEGIN CHUNK <k>" / "END CHUNK <k>" with nothing between if you make no edits),
  4. in the same response, read the next chunk file.
Because every chunk gets its own response, you have the budget to treat EVERY row of every layer; you do not ...
to restrict yourself to a few rows or layers. After the last chunk, send a final message with a short summary
(what you changed and why) and "TOOLS USED: <every tool call>". Do not repeat the edit lines in the final me ...


--- chunk01.txt (start) ---
CHUNK 1/11: conv1 rows 0-7
## conv1 rows 0-7
conv1 r0 scale 0.0635392 bias 0.3771
conv1 r0 u  c0: 3.83 -1.47 0.37 -4.49 2.26 -0.49 0.82 -0.91 -0.55 | c1: 0.14 0.49 -0.50 -7.00 6.61 1.87 -1.1 ...
conv1 r0 q  c0: 4 -1 0 -4 2 0 1 -1 -1 | c1: 0 0 0 -7 7 2 -1 0 1 | c2: -1 -1 -1 -6 5 1 -1 2 1
[...]
```

Figure 8: P2 task description at int4: the parts that differ from P1, and the start of the first chunk. Long lines are cut here.

### A.6 Task description (P3)

The P3 task asks the LLM to quantize the float network itself (Figure 9). It names neither RTN nor the accuracy of a quantized network; it gives only the float accuracy.

```
You are given the complete trained float32 weights of a small convolutional network for CIFAR-10
(10 classes: airplane, automobile, bird, cat, deer, dog, frog, horse, ship, truck). Your task: turn it into a
4-bit weight-quantized network that classifies as accurately as possible, by writing the 4-bit network yourself.

RULES
- No tools: no code execution, no calculator, no files, no internet. Do every computation in your own
  reasoning and write the numbers directly.
- You cannot run the network. There is no data and there are no BatchNorm statistics: the weights in the chunk
  files are all you get.
- The result must be a valid 4-bit network: every layer is stored as integer codes q in [-7, 7] and one positive
  float scale per output row, and the effective weight is w = q * scale[row]. Biases stay float32 exactly as ...
  (you may not change them). The architecture stays as it is. You choose the scale of every row and every code.

NETWORK
Input: 3x32x32 RGB image, normalized per channel with mean (0.4914, 0.4822, 0.4465), std (0.2470, 0.2435, 0. ...
  conv1: Conv2d 3->8,   3x3, padding 1, bias, then ReLU
  conv2: Conv2d 8->16,  3x3, padding 1, bias, then ReLU, then MaxPool 2x2
  conv3: Conv2d 16->32, 3x3, padding 1, bias, then ReLU, then MaxPool 2x2
  conv4: Conv2d 32->48, 3x3, padding 1, bias, then ReLU, then global average pooling
  fc:    Linear 48->10 (logits; prediction = argmax)
The network was trained with BatchNorm after every conv; BatchNorm has been folded into the conv weights and
biases, so the network above is complete as written. On a held-out split the float32 network scores 82.14%.

DUMP FORMAT (in the chunk files)
For every layer and output row r:
  "<layer> r<r> bias <b>"                              the float bias (stays as given)
  "<layer> r<r> w ...", one or more lines              the FLOAT32 weights of that row (3 decimals)
conv layers: values are grouped per input channel; each group is "c<i>:" followed by the 3x3 kernel in
row-major order (position p = 3*kh + kw, 0..8). Long rows are split over several lines.
fc: values are input features 0..47 in order, split over lines of 16 ("i0:" marks the first index on the line).

YOUR ANSWER: ONE LINE PER OUTPUT ROW
  ROW <layer> <r> SCALE <positive float> Q <codes>
<codes> are the integer codes of that row, separated by single spaces, in exactly the order of the weights i ...
dump (conv rows: input channel 0 kernel p=0..8, then input channel 1 kernel, ...; no "c<i>:" labels, no "|";
fc rows: inputs 0..47). A conv row has 9 * (number of input channels) codes (conv1: 27, conv2: 72, conv3: 144,
conv4: 288), an fc row has 48. Example for a conv1 row:  ROW conv1 0 SCALE 0.0635 Q 4 -1 0 -4 2 ... (27 codes).
Write EVERY row of the chunk. A row that is missing, has a wrong number of codes, a code outside [-7, 7] or a
non-positive scale cannot be used and is replaced by plain round-to-nearest quantization of that row.

DELIVERY IN CHUNKS
The dump is NOT in this file. It is delivered in 19 chunk files, read one after another:
  chunk01.txt: conv1 rows 0-7
  chunk02.txt: conv2 rows 0-15
  chunk03.txt: conv3 rows 0-7
  chunk04.txt: conv3 rows 8-15
  chunk05.txt: conv3 rows 16-23
  chunk06.txt: conv3 rows 24-31
  chunk07.txt: conv4 rows 0-3
  chunk08.txt: conv4 rows 4-7
  chunk09.txt: conv4 rows 8-11
  chunk10.txt: conv4 rows 12-15
  chunk11.txt: conv4 rows 16-19
  chunk12.txt: conv4 rows 20-23
  chunk13.txt: conv4 rows 24-27
  chunk14.txt: conv4 rows 28-31
  chunk15.txt: conv4 rows 32-35
  chunk16.txt: conv4 rows 36-39
  chunk17.txt: conv4 rows 40-43
  chunk18.txt: conv4 rows 44-47
  chunk19.txt: fc rows 0-9
Work strictly chunk by chunk. In EACH response:
  1. read exactly one chunk file (the next one in order),
  2. think about that chunk only (you may use everything you have seen so far),
  3. write the ROW lines for ALL rows of that chunk as plain text, between a line "BEGIN CHUNK <k>" and a line
     "END CHUNK <k>",
  4. in the same response, read the next chunk file.
Because every chunk gets its own response, you have the budget to write every row of every layer.
After the last chunk, send a final message with a short summary (how you chose the scales and codes, and why ...
"TOOLS USED: <every tool call>". Do not repeat the ROW lines in the final message.

--- chunk01.txt (start) ---
CHUNK 1/19: conv1 rows 0-7
conv1 r0 bias 0.3771
conv1 r0 w c0: 0.243 -0.093 0.024 -0.285 0.143 -0.031 0.052 -0.058 -0.035 | c1: 0.009 0.031 -0.032 -0.445 0. ...
conv1 r1 bias 0.2284
conv1 r1 w c0: -0.048 -0.170 0.035 -0.008 -0.253 0.042 0.082 0.165 0.148 | c1: 0.064 -0.138 -0.003 0.086 -0. ...
[...]
```

Figure 9: P3 task description, as read by the subagent, and the start of the first chunk file. Long lines are cut here.

### A.7 Task description (P4)

P4 adds the BN statistics to the P3 task and runs from the command line with all tools disabled, so the "No tools" rule is dropped and the chunks arrive as messages (Figure 10). The same text was sent to every LLM.

```
You are given the complete trained float32 weights of a small convolutional network for CIFAR-10
(10 classes: airplane, automobile, bird, cat, deer, dog, frog, horse, ship, truck). Your task: turn it into a
4-bit weight-quantized network that classifies as accurately as possible, by writing the 4-bit network yourself.

RULES
- You cannot run the network. There is no data: the weights and BatchNorm statistics in the chunk
  files are all you get.
- The result must be a valid 4-bit network: every layer is stored as integer codes q in [-7, 7] and one positive
  float scale per output row, and the effective weight is w = q * scale[row]. Biases stay float32 exactly as ...
  (you may not change them). The architecture stays as it is. You choose the scale of every row and every code.

NETWORK
Input: 3x32x32 RGB image, normalized per channel with mean (0.4914, 0.4822, 0.4465), std (0.2470, 0.2435, 0. ...
  conv1: Conv2d 3->8,   3x3, padding 1, bias, then ReLU
  conv2: Conv2d 8->16,  3x3, padding 1, bias, then ReLU, then MaxPool 2x2
  conv3: Conv2d 16->32, 3x3, padding 1, bias, then ReLU, then MaxPool 2x2
  conv4: Conv2d 32->48, 3x3, padding 1, bias, then ReLU, then global average pooling
  fc:    Linear 48->10 (logits; prediction = argmax)
The network was trained with BatchNorm after every conv; BatchNorm has been folded into the conv weights and
biases, so the network above is complete as written. For information, the BatchNorm statistics of the
original network are given per conv output channel (line "<layer> r<r> bn gamma .. beta .. mean .. var .."): the
BatchNorm computed  y = gamma * (x - mean) / sqrt(var + 1e-5) + beta  on the conv output x, where mean and v ...
the running statistics over the training data. The folded weights and bias in the dump are
w_folded = w_raw * gamma / sqrt(var + 1e-5) and b_folded = beta - mean * gamma / sqrt(var + 1e-5); you quant ...
FOLDED weights, the bias stays as given. fc has no BatchNorm. On a held-out split the float32 network scores ...

DUMP FORMAT (in the chunk files)
For every layer and output row r:
  "<layer> r<r> bias <b>"                                  the float bias (stays as given)
  "<layer> r<r> bn gamma <g> beta <b> mean <m> var <v>"     BatchNorm statistics (conv layers only; inf ...
  "<layer> r<r> w ...", one or more lines                  the FLOAT32 weights of that row (3 decimals)
conv layers: values are grouped per input channel; each group is "c<i>:" followed by the 3x3 kernel in
row-major order (position p = 3*kh + kw, 0..8). Long rows are split over several lines.
fc: values are input features 0..47 in order, split over lines of 16 ("i0:" marks the first index on the line).

YOUR ANSWER: ONE LINE PER OUTPUT ROW
  ROW <layer> <r> SCALE <positive float> Q <codes>
<codes> are the integer codes of that row, separated by single spaces, in exactly the order of the weights i ...
dump (conv rows: input channel 0 kernel p=0..8, then input channel 1 kernel, ...; no "c<i>:" labels, no "|";
fc rows: inputs 0..47). A conv row has 9 * (number of input channels) codes (conv1: 27, conv2: 72, conv3: 144,
conv4: 288), an fc row has 48. Example for a conv1 row:  ROW conv1 0 SCALE 0.0635 Q 4 -1 0 -4 2 ... (27 codes).
Write EVERY row of the chunk. A row that is missing, has a wrong number of codes, a code outside [-7, 7] or a
non-positive scale cannot be used and is replaced by plain round-to-nearest quantization of that row.

DELIVERY IN CHUNKS
The weight dump is delivered in 19 messages, one chunk per message, after this one. For EACH chunk message, ...
ROW lines for ALL rows of that chunk as plain text between a line "BEGIN CHUNK <k>" and a line "END CHUNK <k ...
nothing else. You may use everything you have seen so far. After the last chunk I will ask for a short summa ...
you chose the scales and codes. Now reply only with the word READY.

--- first chunk message (start) ---
CHUNK 1/19: conv1 rows 0-7
conv1 r0 bias 0.3771
conv1 r0 bn gamma 0.5747 beta 0.3411 mean -0.0745 var 1.4198
conv1 r0 w c0: 0.243 -0.093 0.024 -0.285 0.143 -0.031 0.052 -0.058 -0.035 | c1: 0.009 0.031 -0.032 -0.445 0. ...
conv1 r1 bias 0.2284
conv1 r1 bn gamma 0.6471 beta 0.2579 mean 0.0815 var 3.2010
[...]
```

Figure 10: P4 task description as sent from the command line, and the start of the first chunk message. Long lines are cut here.